\pdfoutput=1

\documentclass[11pt]{article}

\usepackage[]{acl}
\usepackage{hyperref}
\usepackage{nicematrix}
\usepackage{cleveref}

\usepackage{times}
\usepackage{latexsym}

\usepackage[T1]{fontenc}

\usepackage[utf8]{inputenc}

\usepackage{microtype}
\usepackage{booktabs}
\usepackage{adjustbox}
\usepackage{multirow}
\usepackage{amsfonts}
\usepackage{enumerate}
\usepackage{siunitx}
\usepackage[xcolor]{colortbl}
\DeclareUnicodeCharacter{2581}{{\rmfamily\textbf{\_}}}
\usepackage{subfig}
\usepackage{graphicx}
\usepackage{relsize}
\usepackage{placeins}

\DeclareUnicodeCharacter{1EA1}{\`{a}} %
\DeclareUnicodeCharacter{01A1}{\^{e}} %
\DeclareUnicodeCharacter{1EF9}{\~{y}} %
\DeclareUnicodeCharacter{01B0}{\^o}

\title{
A Comparison of Language Modeling and Translation as Multilingual Pretraining Objectives
}

\author{Zihao Li,$^{1}$~Shaoxiong Ji,$^*$$^{1}$~Timothee Mickus,$^*$$^{1}$~ Vincent Segonne,$^{2}$ \and Jörg Tiedemann$^{1}$ \\
$^{1}$University of Helsinki \quad  $^{2}$ Université Bretagne Sud
\\
\texttt{firstname.lastname@\{$^{1}$helsinki.fi,~$^{2}$univ-ubs.fr\}}
}

\begin{document}
\maketitle
\def\thefootnote{*}\footnotetext{Equal contribution and corresponding authors.}\def\thefootnote{\arabic{footnote}}
\begin{abstract}

Pretrained language models (PLMs) display impressive performances and have captured the attention of the NLP community.
Establishing best practices in pretraining has, therefore, become a major focus of NLP research, especially since insights gained from monolingual English models may not necessarily apply to more complex multilingual models.
One significant caveat of the current state of the art is that different works are rarely comparable: they often discuss different parameter counts, training data, and evaluation methodology.

This paper proposes a comparison of multilingual pretraining objectives in a controlled methodological environment. We ensure that training data and model architectures are comparable, and discuss the downstream performances across 6 languages that we observe in probing and fine-tuning scenarios.
We make two key observations: (1) the architecture dictates which pretraining objective is optimal; (2) multilingual translation is a very effective pretraining objective under the right conditions.
We make our code, data, and model weights available at \texttt{\url{https://github.com/Helsinki-NLP/lm-vs-mt}}.

\end{abstract}

\section{Introduction}
\label{sec:introduction}

The release of BERT \citep{devlin-etal-2019-bert} has marked a paradigm shift in the NLP landscape and has ushered in a thorough investment of the NLP research community in developing large language models that can readily be adapted to novel situations.
The design, training, and evaluation of these models has become a significant enterprise of its own.

In recent years, that sustained interest has shifted also to encompass multilingual models \citep[e.g.,][]{muennighoff2022crosslingual,alves2024tower}.
There is considerable variation as to how such models are trained: 
For instance, some rely on datasets comprising multiple languages without explicit cross-lingual supervision (e.g., \citealp{liu-etal-2020-multilingual-denoising}), and some use explicit supervision \citep{xue-etal-2021-mt5}.
One complication that arises from this blossoming field of study is that much of the work being carried out is not directly comparable beyond the raw performances on some well-established benchmark, a procedure which may well be flawed \citep{gorman-bedrick-2019-need}.
Avoiding apples-to-oranges comparison requires a methodical approach in strictly comparable circumstances, which is the stance we adopt in this paper.

In short, we focus on two variables---model architecture and pretraining objectives---and set out to train five models in strictly comparable conditions and compare their monolingual performances in three downstream applications: sentiment analysis, named entity recognition, and POS-tagging. 
The scope of our study spans from encoder-decoder machine translation models, to decoder-only causal language models and encoder-only BERT-like masked language models. We categorize them into \textbf{double-stacks (encoder-decoder)} and \textbf{single-stacks (encoder-only or decoder-only)} models.
We intend to answer two research questions:
\begin{enumerate}[(i)]
    \item Does the explicit cross-lingual training signal of translation objectives foster better downstream performances in monolingual tasks?
    \item Is the optimal choice of architecture independent of the training objective?
\end{enumerate}
There are \textit{a prima facie} reasons to favor either answers to both of these questions.
For instance, the success of multilingual pretrained language models (LM) on cross-lingual tasks has been underscored repeatedly \citep[e.g.,]{wu2019beto}, yet explicit alignments such as linear mapping~\citep{wang2019cross} and L2 alignment~\citep{cao2020multilingual} between source and target languages do not necessarily improve the quality of cross-lingual representations \citep{wu2020explicit}.

Our experiments provide tentative evidence that
insofar as a BART denoising autoencoder architecture is concerned, models pretrained with a translation objective consistently outperform those trained with a denoising objective.
However, for single-stack transformers, we observe causal language models to perform well in probing and masked language models to generally outperform translation and causal objectives when fine-tuned on downstream tasks.
This leads us to conjecture that the optimal pretraining objective depends on the architecture. Furthermore, the best downstream results we observe appear to stem from a machine-translation system, highlighting that MT encoder-decoder systems might constitute an understudied but potentially very impactful type of pretrained model.

\section{Methods and Settings}
\label{sec:settings}

We start our inquiry by adopting a principled stance: We train strictly comparable models with MT and LM objectives before contrasting their performances on monolingual tasks.

\paragraph{Models and objectives.}
To allow a systematic evaluation, we train models with various neural network architectures and learning objectives.
All models are based on the transformer architecture \citep{NIPS2017_3f5ee243} and implemented in \texttt{fairseq} \citep{ott-etal-2019-fairseq}. 
We consider both double-stacks (encoder-decoder) and single-stacks (encoder-only or decoder-only) models. 

The two double-stack models are variants of the BART architecture of \citep{lewis-etal-2020-bart};
they are trained either on a straightforward machine translation (MT) objective, using language tokens to distinguish the source, or on the original denoising auto-encoder objective of \citeauthor{lewis-etal-2020-bart}. We refer to these two models as \textbf{2-LM} and \textbf{2-MT} respectively.

We also consider three single-stack models:
(i) an encoder-only model trained on the masked language modeling objective (\textbf{MLM}) of \citet{devlin-etal-2019-bert};
(ii) an autoregressive causal language model (\textbf{CLM}), similar to \citet{radford2019language}; 
and (iii) an autoregressive model trained to generate a sentence, followed by its translation in the language specified by a given control token, known as a translation language model (\textbf{TLM}) as proposed by \citet{NEURIPS2019_c04c19c2}.\footnote{
    In this work, we only focus on the causal variant of TLM proposed by \citeauthor{NEURIPS2019_c04c19c2}.
}
We provide an example datapoint for each pretraining objective in \Cref{tab:objectives}, \Cref{adx:objectives}.

\paragraph{Pretraining conditions.}
Our core focus is on guaranteeing comparable conditions across the different pretraining objectives we consider.
This entails that our datasets need to be doubly structured: both in documents for CLM pretraining; and as aligned bitexts for MT pretraining.
Two datasets broadly match these criteria: the UNPC~\citep{ziemski-etal-2016-united} and OpenSubtitles~(OpSub; \citealp{tiedemann2012parallel}) corpora. 
The choice also narrows down the languages considered in this study: we take the set of languages present in both resources, namely the six languages in UNPC: Arabic (AR), Chinese (ZH), English (EN), French (FR), Russian (RU), and Spanish (ES).

To guarantee that models are trained on the same data, whenever a document is available in multiple languages, we greedily assign it to the least represented language pair thus far and discard all other possible language pairs where it could have contributed; we then discard documents which cannot be used as bitexts. 
This ensures that all documents are used exactly once for both document-level and bitext-level pretraining objectives. %
Dataset statistics are shown
in \Cref{tab:pretraining-dataset}, \Cref{adx:downstream-dataset-details}.

To ensure a fair comparison, we control key variables, including tokenization (100k BPE pieces; \citealp{sennrich-etal-2016-neural}), number of transformer layers (12), hidden dimensions (512), attention heads (8), and feedforward layer dimensions (2048). 
We perform 600k steps of updates,\footnote{
    Improvements in cross-entropy over the validation set were always marginal after this stage. 
} using the largest batch size that fits into the GPU memory, deploy distributed training to make a global batch size of 4096, and apply the Adam optimizer \citep{kingma2017adam}.
Owing to the computational requirements, we only train one seed for each of the five types of models considered.

\paragraph{Downstream evaluation.}
The evaluations encompassed both sequence-level and token-level classification tasks using datasets tailored for sentiment analysis (SA), named entity recognition (NER), part-of-speech (POS) tagging, and natural language inference (NLI).

For SA, we utilized the Amazon review dataset~\citep{hou2024bridging} in English, Spanish, French, and Chinese. RuReviews~\citep{Smetanin-SA-2019} for Russian, and \texttt{ar\_res\_reviews}~\citep{elsahar2015building} for Arabic. %
While the datasets for most languages were pre-split, \texttt{ar\_res\_reviews} required manual division into training, validation, and testing sets, using an 8:1:1 ratio. %

For NER, we model the problem as an entity span extraction using a BIO scheme. 
In practice, we classify tokens into three basic categories: Beginning of an entity (B), Inside an entity (I), or Outside any entity (O). We use the MultiCoNER v2 dataset~\citep{fetahu-etal-2023-multiconer} for English, Spanish, French, and Chinese, MultiCoNER v1~\citep{malmasi-etal-2022-multiconer} for Russian and the AQMAR Wikipedia NER corpus~\citep{mohit-etal-2012-recall} for Arabic. Simplifying the NER task to these fundamental categories allows us to focus more on assessing the basic entity recognition capabilities of the models without the additional complexity of differentiating numerous entity types, which can vary significantly between languages and datasets.

For POS tagging, we utilized the Universal Dependencies (UD) 2.0 datasets~\citep{nivre2020universal}, selecting specific corpora tailored to each language to ensure both linguistic diversity and relevance. 
We select multiple UD treebanks per language, such that
each language dataset comprises approximately 160,000 tokens, which are then split into training, validation, and testing segments with an 8:1:1 ratio. 

For NLI, we employed the XNLI dataset~\citep{conneau-etal-2018-xnli} for the six languages. The XNLI dataset consists of sentence pairs translated from the MultiNLI dataset~\citep{williams-etal-2018-broad} into 15 languages, providing consistent annotations across languages. The task focuses on classifying the relationship between pairs of sentences into one of three categories: Entailment, Contradiction, or Neutral. Unlike the original cross-lingual design of XNLI, we conducted monolingual experiments for each language to evaluate the performance of our models individually in each linguistic context.

Supplementary details regarding data preprocessing for downstream experiments are available in \Cref{adx:downstream-dataset-details}.

We evaluate the performances of the encoder output representations for the 2-MT and 2-LM models and of the last hidden representation before the vocabulary projection for the single-stack models.

The evaluation of the models involves two distinct experimental approaches to test the performance: probing and fine-tuning. In the probing experiments, only the parameters of the classification heads are adjusted. This method primarily tests the raw capability of the pre-trained models' embeddings to adapt to specific tasks with minimal parameter changes, preserving the underlying pre-trained network structure.
Conversely, in the fine-tuning experiments, all parameters of the models are adjusted. This approach allows the entire model to adapt to the specifics of the task, potentially leading to higher performance at the cost of significantly altering the pre-trained weights.

For both experimental approaches, each model is trained for 10 epochs to ensure sufficient learning without overfitting. We optimize parameters with AdamW~\citep{loshchilov2017decoupled}, with a constant learning rate of 0.0001 across all tasks and models. 
This setup was chosen to standardize the training process, providing a fair basis for comparing the performance outcomes across different models and tasks.
We reproduce probing and fine-tuning for 5 seeds to ensure stability.

\section{Results}

\begin{table*}[t]
		\centering
        \subfloat[\label{tab:BART-probing} Probing]{\resizebox{0.49\linewidth}{!}{
			\setlength{\tabcolsep}{3pt}
			\begin{tabular}{l@{{~~}}l c@{{\smaller$\pm$}}>{\smaller}c@{{~~}} c@{{\smaller$\pm$}}>{\smaller}c@{{~~}} c@{{\smaller$\pm$}}>{\smaller}c@{{~~}} c@{{\smaller$\pm$}}>{\smaller}c@{{~~}} c@{{\smaller$\pm$}}>{\smaller}c@{{~~}} c@{{\smaller$\pm$}}>{\smaller}c}
				\toprule
                
				\multicolumn{2}{c}{\multirow{2}{*}{\textbf{Setup}}} & 
				\multicolumn{12}{c}{\textbf{Languages}} \\
				& &	\multicolumn{2}{c}{\textbf{EN}}	&	\multicolumn{2}{c}{\textbf{ES}}	&	\multicolumn{2}{c}{\textbf{FR}}	&	\multicolumn{2}{c}{\textbf{ZH}}	&	\multicolumn{2}{c}{\textbf{RU}}	&	\multicolumn{2}{c}{\textbf{AR}}	\\

				\midrule
				\multirow{2}{*}{\rotatebox{90}{\textbf{SA}}}
				& 2-LM &	
                42.86 & 0.86 & 42.80	& 0.69 &  43.00 & 0.60 &  40.41 & 1.02 & 	65.83 & 0.70 &  70.88 & 1.62	\\
				& 2-MT &	
                \textbf{46.71} & 0.88 &  \textbf{46.64} & 0.55 &  \textbf{46.10} & 0.43 &  \textbf{43.74} & 0.65 &  \textbf{68.79} & 0.42 &  \textbf{73.77} & 0.97 \\
				\midrule
				
				\multirow{2}{*}{\rotatebox{90}{\textbf{NER}}} 
				& 2-LM &	
                82.69 & 0.09 & 
                84.74 & 0.07 & 
                82.80 & 0.06 & 
                78.88 & 0.25 & 
                77.93 & 0.15 & 
                85.28 & 0.22 \\
				& 2-MT &	
                \textbf{89.47} & 0.06  & 
                \textbf{90.54} & 0.04  & 
                \textbf{89.41} & 0.10  & 
                \textbf{88.78} & 0.09 & 
                \textbf{83.39} & 0.22  & 
                \textbf{89.70} & 0.18  \\
				\midrule
				
				\multirow{2}{*}{\rotatebox{90}{\textbf{POS}}} 
				& 2-LM &
                78.85 & 0.29  & 78.12 & 0.25  & 81.57 & 0.32  & 66.09 & 0.25  & 77.93 & 0.12  & 47.68 & 0.10  \\
				& 2-MT &	
                \textbf{92.22} & 0.14  & \textbf{90.59} & 0.20  & \textbf{95.39} & 0.10  & \textbf{75.87} & 0.17 & \textbf{93.20} & 0.08  & \textbf{61.84} & 0.24  \\
                \midrule

                \multirow{2}{*}{\rotatebox{90}{\textbf{NLI}}} 
                & 2-LM &
                48.56 & 0.01 & 49.31 & 0.01 & 48.33 & 0.01 & 38.81 & 0.01 & 48.34 & 0.01 & 45.11 & 0.01 \\
                & 2-MT &
                \textbf{60.50} & 0.01 & \textbf{59.56} & 0.01 & \textbf{59.00} & 0.01 & \textbf{59.01} & 0.01 & \textbf{59.83} & 0.01 & \textbf{59.58} & 0.01 \\
                
				\bottomrule
			\end{tabular}
		}}
		\subfloat[\label{tab:BART-finetuning}Fine-tuning]{\resizebox{0.49\linewidth}{!}{
			\setlength{\tabcolsep}{3pt}
                \begin{tabular}{l@{{~~}}l c@{{\smaller$\pm$}}>{\smaller}c@{{~~}} c@{{\smaller$\pm$}}>{\smaller}c@{{~~}} c@{{\smaller$\pm$}}>{\smaller}c@{{~~}} c@{{\smaller$\pm$}}>{\smaller}c@{{~~}} c@{{\smaller$\pm$}}>{\smaller}c@{{~~}} c@{{\smaller$\pm$}}>{\smaller}c}
				\toprule
				\multicolumn{2}{c}{\multirow{2}{*}{\textbf{Setup}}} & 
				\multicolumn{12}{c}{\textbf{Languages}} \\
				\multicolumn{2}{c}{\multirow{2}{*}{}} &	\multicolumn{2}{c}{\textbf{EN}}	&	\multicolumn{2}{c}{\textbf{ES}}	&	\multicolumn{2}{c}{\textbf{FR}}	&	\multicolumn{2}{c}{\textbf{ZH}}	&	\multicolumn{2}{c}{\textbf{RU}}	&	\multicolumn{2}{c}{\textbf{AR}}	\\
				\midrule
				\multirow{2}{*}{\rotatebox{90}{\textbf{SA}}}
				& 2-LM 
                &	52.26 & 0.55  & 52.89 & 0.69  & 52.99 & 0.59  & 48.64 & 0.36 & 
                73.89 & 0.43  & 79.74 & 1.36  \\
				& 2-MT &	\textbf{54.76} & 0.58  & \textbf{55.56} & 0.49  & \textbf{54.75} & 0.42  & \textbf{50.55} & 0.68 & \textbf{74.77} & 0.50  & \textbf{81.49} & 1.49  \\
				\midrule
				\multirow{2}{*}{\rotatebox{90}{\textbf{NER}}}
				& 2-LM &
                91.13 & 0.12  & 91.82 & 0.21  & 91.58 & 0.10  & 92.30 & 0.10 & 
                85.34 & 0.39  & 89.05 & 0.13  \\
				& 2-MT &	\textbf{93.46} & 0.09  & \textbf{94.22} & 0.09  & \textbf{93.84} & 0.04  & \textbf{93.75} & 0.32 & \textbf{89.07} & 0.11  & \textbf{93.26} & 0.15  \\
				\midrule
				\multirow{2}{*}{\rotatebox{90}{\textbf{POS}}} 
				& 2-LM &	
                92.42 & 0.28  & 90.41 & 0.16 & 
                95.21 & 0.13  & 82.30 & 0.48 & 
                95.36 & 0.20  & 69.57 & 0.24  \\
				& 2-MT &	\textbf{95.98} & 0.08  & \textbf{94.29} & 0.05  & \textbf{98.05} & 0.17  & \textbf{90.18} & 0.15 & \textbf{97.00} & 0.07  & \textbf{74.47} & 0.08  \\
                \midrule
                \multirow{2}{*}{\rotatebox{90}{\textbf{NLI}}} 
                & 2-LM &
                57.76 & 0.01 & 57.87 & 0.01 & 56.77 & 0.01 & 48.05 & 0.01 & 56.43 & 0.01 & 0.5377 & 0.01 \\
                & 2-MT &
                \textbf{61.96} & 0.01 & \textbf{61.71} & 0.01 & \textbf{60.09} & 0.01 & \textbf{53.72} & 0.01 & \textbf{59.00} & 0.01 & \textbf{0.569}3 & 0.01 \\
				\bottomrule
			\end{tabular}
		}}
		\caption{Accuracy ($\times$ 100) of double-stack models ($\pm$ s.d. over 5 runs).}
	\end{table*}

\paragraph{Double-stack models.}
We first compare the performance of 2-LM and 2-MT across several key language processing tasks including SA, NER, POS tagging, and NLI. 
Results are shown in \Cref{tab:BART-probing,tab:BART-finetuning}.
The pretraining objectives play a significant role in shaping the models' effectiveness. 
Specifically, 2-MT, which is pretrained with a machine translation objective, consistently outperforms 2-LM, which utilizes a denoising objective. This pattern is consistent across all languages tested after fine-tuning as well as probing.

\begin{table*}[t]
        \centering
        \subfloat[\label{tab:xLm-probing}Probing]{\resizebox{0.48\linewidth}{!}{
			\setlength{\tabcolsep}{3pt}
			\begin{tabular}{l@{{~~}}l c@{{\smaller$\pm$}}>{\smaller}c@{{~~}} c@{{\smaller$\pm$}}>{\smaller}c@{{~~}} c@{{\smaller$\pm$}}>{\smaller}c@{{~~}} c@{{\smaller$\pm$}}>{\smaller}c@{{~~}} c@{{\smaller$\pm$}}>{\smaller}c@{{~~}} c@{{\smaller$\pm$}}>{\smaller}c}
				\toprule
                
				\multicolumn{2}{c}{\multirow{2}{*}{\textbf{Setup}}} & 
				\multicolumn{12}{c}{\textbf{Languages}} \\
				& &	\multicolumn{2}{c}{\textbf{EN}}	&	\multicolumn{2}{c}{\textbf{ES}}	&	\multicolumn{2}{c}{\textbf{FR}}	&	\multicolumn{2}{c}{\textbf{ZH}}	&	\multicolumn{2}{c}{\textbf{RU}}	&	\multicolumn{2}{c}{\textbf{AR}}	\\

            \midrule
            \multirow{3}{*}{\rotatebox{90}{\textbf{SA}}}
            & CLM 
            & \textbf{35.14} & 0.92  
            & \textbf{35.66} & 1.10 
            & \textbf{34.14} & 1.63  
            & \textbf{33.62} & 0.83 
            & \textbf{57.57} & 1.11   
            & \textbf{67.71}	& 2.24   \\
            & MLM 
            & 34.26 & 1.34  
            & 34.82 & 1.58 
            & 33.90 & 1.12   
            & 32.52 & 1.65 
            & 54.55 & 1.86   
            & 65.94 & 3.30   \\
            & TLM 
            & 29.68 & 2.22  
            & 32.20	& 3.07  
            & 32.26 & 2.34 
            & 29.88 & 4.17 
            & 56.45 & 1.81   
            & 64.45	& 1.81   \\
            \midrule
            
            \multirow{3}{*}{\rotatebox{90}{\textbf{NER}}} 
            & CLM 
            & \textbf{80.27} & 0.12  
            & \textbf{82.59} & 0.06   
            & \textbf{80.38} & 0.12   
            & \textbf{77.92} & 0.28 
            & \textbf{76.39} & 0.03   
            & 84.17 & 0.08  \\
            & MLM 
            & 78.77 & 0.02  
            & 81.61 & 0.00   
            & 79.11 & 0.01  
            & 70.67 & 0.10 
            & 76.34 & 0.01  
            & \textbf{84.29} & 0.00  \\
            & TLM 
            & 79.10 & 0.06   
            & 81.94 & 0.13   
            & 79.56 & 0.14   
            & 77.26 & 0.24 
            & \textbf{76.39} & 0.02  
            & 84.26 & 0.02   \\
            \midrule
            
            \multirow{3}{*}{\rotatebox{90}{\textbf{POS}}} 
            & CLM 
            & \textbf{69.06} & 0.38   
            & \textbf{70.32} & 0.50   
            & \textbf{76.67} & 0.46   
            & \textbf{51.40} & 0.47 
            & \textbf{59.64} & 0.62   
            & \textbf{43.49} & 0.40  \\
            & MLM 
            & 37.92 & 0.61   
            & 44.26 & 0.11   
            & 46.89 & 0.32   
            & 31.16 & 0.21 
            & 34.62 & 0.16   
            & 34.71 & 0.94  \\
            & TLM 
            & 62.96 & 1.02   
            & 62.08 & 1.99  
            & 63.89 & 1.06   
            & 50.46 & 0.53 
            & 54.27 & 0.87   
            & 40.94 & 1.16  \\
            \midrule

            \multirow{3}{*}{\rotatebox{90}{\textbf{NLI}}} 
            & CLM 
            & 42.32 & 0.02 
            & 42.99 & 0.01 
            & \textbf{43.43} & 0.02 
            & 40.55 & 0.02 
            & 40.06 & 0.02 
            & 41.99 & 0.01 \\
            & MLM 
            & \textbf{45.64} & 0.02 
            & \textbf{44.49} & 0.01 
            & 43.11 & 0.02 
            & \textbf{42.80} & 0.01 
            & \textbf{43.16} & 0.01 
            & \textbf{43.55} & 0.01 \\
            & TLM 
            & 38.36 & 0.02 
            & 41.95 & 0.02 
            & 41.89 & 0.01 
            & 38.93 & 0.04 
            & 41.20 & 0.02 
            & 39.50 & 0.02 \\
            
            \bottomrule
            \end{tabular}
        }}
        \subfloat[\label{tab:xLm-finetuning}Fine-tuning]{\resizebox{0.5\linewidth}{!}{
			\setlength{\tabcolsep}{3pt}
			\begin{tabular}{l@{{~~}}l c@{{\smaller$\pm$}}>{\smaller}r@{{~~}} c@{{\smaller$\pm$}}>{\smaller}r@{{~~}} c@{{\smaller$\pm$}}>{\smaller}r@{{~~}} c@{{\smaller$\pm$}}>{\smaller}r@{{~~}} c@{{\smaller$\pm$}}>{\smaller}r@{{~~}} c@{{\smaller$\pm$}}>{\smaller}r}
				\toprule
                
				\multicolumn{2}{c}{\multirow{2}{*}{\textbf{Setup}}} & 
				\multicolumn{12}{c}{\textbf{Languages}} \\
				& &	\multicolumn{2}{c}{\textbf{EN}}	&	\multicolumn{2}{c}{\textbf{ES}}	&	\multicolumn{2}{c}{\textbf{FR}}	&	\multicolumn{2}{c}{\textbf{ZH}}	&	\multicolumn{2}{c}{\textbf{RU}}	&	\multicolumn{2}{c}{\textbf{AR}}	\\            
            \midrule
            \multirow{3}{*}{\rotatebox{90}{\textbf{SA}}}
            & CLM 
            & \textbf{55.23} & 0.72   
            & 47.81 & 15.55   
            & 54.84 & 0.62   
            & 51.18 & 0.94 
            & 75.07 & 0.21  
            & 66.18 & 21.74  \\
            & MLM 
            & 55.22 & 0.92   
            & 55.67 & 1.77   
            & 54.08 & 2.43  
            & 51.00 & 1.07
            & 74.53 & 1.36 
            & \textbf{75.00} & 3.48   \\
            & TLM 
            & 55.14 & 0.92   
            & \textbf{55.84} & 0.59   
            & \textbf{55.22} & 0.98  
            & \textbf{51.46} & 0.53 
            & \textbf{75.31} & 0.57  
            & 72.75 & 2.25  \\
            \midrule
            
            \multirow{3}{*}{\rotatebox{90}{\textbf{NER}}} 
            & CLM 
            & 89.91 & 0.33   
            & 91.42 & 0.15   
            & 90.65 & 0.17   
            & 89.97 & 0.14 
            & 83.20 & 0.31   
            & \textbf{87.50} & 2.22   \\
            & MLM 
            & \textbf{93.31} & 0.57   
            & \textbf{93.93} & 0.60   
            & \textbf{93.67} & 0.30   
            & \textbf{92.99} & 0.99 
            & \textbf{87.49} & 0.78  
            & 85.78 & 3.30  \\
            & TLM 
            & 89.88 & 0.06  
            & 91.45 & 0.25  
            & 90.49 & 0.23  
            & 90.10 & 0.14 
            & 83.76 & 0.63   
            & 84.29 & 0.00  \\
            \midrule
            
            \multirow{3}{*}{\rotatebox{90}{\textbf{POS}}} 
            & CLM 
            & 91.72 & 0.14   
            & 90.51 & 0.13   
            & 95.75 & 0.10   
            & 78.61 & 0.31
            & 85.50 & 0.15  
            & 57.43 & 1.63   \\
            & MLM 
            & \textbf{96.00} & 0.15 
            & \textbf{94.45} & 0.13   
            & \textbf{97.94} & 0.20   
            & \textbf{89.96} & 0.71 
            & \textbf{96.69} & 0.13   
            & \textbf{74.35} & 0.53   \\
            & TLM 
            & 91.68 & 0.19   
            & 90.38 & 0.20   
            & 86.99 & 19.40   
            & 78.50 & 0.52 
            & 85.71 & 0.18   
            & 59.11 & 0.50  \\
            \midrule

            \multirow{3}{*}{\rotatebox{90}{\textbf{NLI}}} 
            & CLM 
            & 48.84 & 0.14 
            & 56.46 & 0.03 
            & \textbf{55.45} & 0.03 
            & \textbf{49.70} & 0.06 
            & 55.23 & 0.02 
            & 49.02 & 0.07 \\
            & MLM 
            & \textbf{59.41} & 0.01 
            & \textbf{57.54} & 0.04 
            & 55.04 & 0.06 
            & 47.96 & 0.03 
            & \textbf{57.80} & 0.01 
            & \textbf{53.60} & 0.01 \\
            & TLM 
            & 49.76 & 0.10 
            & 52.12 & 0.11 
            & 54.20 & 0.10 
            & 49.03 & 0.04 
            & 53.60 & 0.04 
            & 44.39 & 0.10 \\
            \bottomrule
            \end{tabular}
        }}
        
        \caption{Accuracy ($\times$ 100) of single-stack models ($\pm$ s.d. over 5 runs).}
\end{table*}

\paragraph{Single-stack models.}
Turning to the single-stack models (CLM, MLM, TLM), we find a somewhat more complex picture.
In a probing context (cf. \Cref{tab:xLm-probing}), we find the CLM to be almost always the most effective, except for NLI in five languages and NER in Arabic, where it performs slightly less favorably compared to the MLM.
As for fine-tuning (\Cref{tab:xLm-finetuning}), while the MLM generally ranks first on all POS, NER, and NLI datasets, the TLM is usually effective for SA.\footnote{
    However, remark that unlike with the BART-based models, SA results are not stable when we shift metrics from accuracy to F1 (see \Cref{tab:F1-Probing,tab:F1-Finetuning} in \Cref{adx:benchmarks}). 
    The difference in F1 between the top two models is often $\leq 0.01$, making it difficult to ascertain that one model strictly dominates.
}

\paragraph{Discussion.}
A first global observation that we can make for these results is that single-stack and double-stack models appear to behave differently. 
While the MT objective yields the highest performances for BART-type models, the downstream performances of the TLM do not really stand out compared to the CLM in probing and the MLM in fine-tuning scenarios.
It is important to note that the performances stem at least in part from the architecture itself:
2-MT and 2-LM both consistently outperform all single-stack models in probing.
However, it is crucial to acknowledge the limitations of our study, as we only conducted one pretraining round for all the objectives. Hence, this evidence should be interpreted as tentative at best.

Fine-tuning also tends to minimize the difference between single-stack and double-stack models---which suggests that the higher quality of double-stack representations could be an artifact of training limitations.
Moreover, the relative ranks of the three single-stack models fluctuate much more than what we see for the double-stack models, owing to no little extent to the oftentimes momentous variation across seeds for single-stack models.
We therefore conjecture that while a translation objective can yield a clear training signal towards semantically informed representations, this comes with two caveats: first, the signal can only be leveraged with dedicated separate modeling of source and target (viz. double-stack models); second, this advantage is much less consequential when fine-tuning.

\section{Related works} \label{sec:sota}

Multilingual foundation models have flourished in recent years (a.o., \citealp{NEURIPS2019_c04c19c2,liu-etal-2020-multilingual-denoising,xue-etal-2021-mt5, kale-etal-2021-nmt5, fang2021filter, chi-etal-2021-mt6,alves2024tower,üstün2024aya}), and with them so have studies of their representations (\citealp{conneau-etal-2020-emerging,siddhant2020evaluating,Choudhury_Deshpande_2021,fierro-sogaard-2022-factual,haemmerl-etal-2023-exploring} a.o.).
All of these works, however, fail to control for some of the most crucial factors, such as ensuring that all models are trained on comparable amounts of data.

This work is specifically related to \citet{NEURIPS2019_c04c19c2}, which also compares MLM, CLM, and TLM but does not normalize the training data.
Another point of comparison is \citet{ji-etal-2024-machine-translation}, which studies the impact of MT continued pretraining in BART on cross-lingual downstream tasks. 
Monolingual evaluation of multilingual systems has also been broached a.o. by \citet{rust-etal-2021-good}. 

\section{Conclusion}
\label{sec:conclusion}

This paper conducts an empirical study of how pretraining conditions of multilingual models impact downstream performances in probing and fine-tuning scenarios.
Despite the inherent limitations that stem from our stringent data requirements, our experiments offer a novel perspective that highlights directions for future inquiry into how multilingual foundation models ought to be pretrained. 
We observe
that double-stack BART-based models fare much better than single-stack models in probing scenarios, but the difference is overall less clear when it comes to fine-tuning.
We also find some tentative evidence that translation objectives can be highly effective for model pretraining in precise circumstances: 
Namely, the most effective model %
on downstream tasks 
among those we experimented with
is an MT-pretrained BART-like model, which outperforms both a more traditional denoising objective for BART as well as decoder-only CLM and encoder-only MLM models.
This would suggest that translation can serve as a powerful pretraining objective, although it is currently under-explored.\footnote{
    There are reasonable objections against using MT models as pretrained multilingual foundation models---namely, unlike auto-regressive causal language models, their generation capabilities are strictly tied to translation, thereby requiring some degree of multilingualism from end-users.
}

Another crucial aspect of our study is that we present strictly comparable models, trained on comparable data, with comparable parameter counts and unified implementations.
While this entails some limitations, especially with regard to the scale of models and data used, we nonetheless believe that a strict comparison can help discriminate between the various factors at play in other works. 
Here, we find clear evidence that CLM pretraining objectives, such as those used in GPT, outperform MLM-based models, such as BERT, in probing scenarios;
we are also able to isolate and highlight how the optimal choice of pretraining objective is contingent on the architecture being employed.

For future work, we recommend exploring \emph{multitask learning} during pretraining by combining objectives like translation, denoising, and language modeling; in such cases, models could harness the strengths of each task to become more robust and versatile. Additionally, investigating \emph{training-free} evaluation methods can offer insights into a model's inherent capabilities without the variability introduced by fine-tuning.

\section*{Acknowledgments}

We thank Alessandro Raganato and our colleagues at the Helsinki-NLP group for useful discussions throughout this project, as well as the three anonymous reviewers for their comments. 

This project has received funding from the European Union’s Horizon Europe research and innovation programme under Grant agreement No 101070350 and from UK Research and Innovation (UKRI) under the UK government’s Horizon Europe funding guarantee [grant number 10052546], and partially funded by the French National Research Agency [grant ANR-23-IAS1-0001]. The contents of this publication are the sole responsibility of its authors and do not necessarily reflect the opinion of the European Union.

The authors wish to thank CSC-IT Center for Science, Finland, for the generous computational resources on the Puhti supercomputer and LUMI supercomputer through the LUMI extreme scale access (MOOMIN and LumiNMT). Some of the experiments were performed using the Jean Zay and Adastra clusters from GENCI-IDRIS [grant 2022 A0131013801].

\section*{Limitations}

This study employs models that are not large in terms of parameters in the era of large language models. Such a constraint potentially hinders the generalizability of our results to much larger architectures that are capable of handling a broader array of linguistic nuances.
Furthermore, our study focuses on a small selected group of languages and specific NLP tasks. This focus might limit the applicability of our findings to other linguistic contexts or more complex real-world applications where diverse language phenomena or different task demands play a crucial role.

Another limitation is our reliance on specific corpora. The datasets utilized, while valuable, represent a potential source of selection bias. They may not fully encompass the vast diversity of global language use, thus skewing the model training and evaluation. 
Such a bias could affect the robustness and effectiveness of the pretrained models when applied to languages that are not well-represented in the training data.

\bibliography{anthology,lm-vs-mt}

\begin{thebibliography}{52}
\expandafter\ifx\csname natexlab\endcsname\relax\def\natexlab#1{#1}\fi

\bibitem[{{AllSet Learning}(2023)}]{ChineseGrammarWiki}
{AllSet Learning}. 2023.
\newblock \href {https://resources.allsetlearning.com/chinese/grammar/}
  {Chinese grammar wiki}.

\bibitem[{Alves et~al.(2024)Alves, Pombal, Guerreiro, Martins, Alves, Farajian,
  Peters, Rei, Fernandes, Agrawal, Colombo, de~Souza, and
  Martins}]{alves2024tower}
Duarte~M. Alves, José Pombal, Nuno~M. Guerreiro, Pedro~H. Martins, João
  Alves, Amin Farajian, Ben Peters, Ricardo Rei, Patrick Fernandes, Sweta
  Agrawal, Pierre Colombo, José G.~C. de~Souza, and André F.~T. Martins.
  2024.
\newblock \href {http://arxiv.org/abs/2402.17733} {Tower: An open multilingual
  large language model for translation-related tasks}.

\bibitem[{Cao et~al.(2020)Cao, Kitaev, and Klein}]{cao2020multilingual}
Steven Cao, Nikita Kitaev, and Dan Klein. 2020.
\newblock Multilingual alignment of contextual word representations.
\newblock In \emph{International Conference on Learning Representations}.

\bibitem[{Chi et~al.(2021)Chi, Dong, Ma, Huang, Singhal, Mao, Huang, Song, and
  Wei}]{chi-etal-2021-mt6}
Zewen Chi, Li~Dong, Shuming Ma, Shaohan Huang, Saksham Singhal, Xian-Ling Mao,
  Heyan Huang, Xia Song, and Furu Wei. 2021.
\newblock \href {https://doi.org/10.18653/v1/2021.emnlp-main.125} {m{T}6:
  Multilingual pretrained text-to-text transformer with translation pairs}.
\newblock In \emph{Proceedings of the 2021 Conference on Empirical Methods in
  Natural Language Processing}, pages 1671--1683, Online and Punta Cana,
  Dominican Republic. Association for Computational Linguistics.

\bibitem[{Choudhury and Deshpande(2021)}]{Choudhury_Deshpande_2021}
Monojit Choudhury and Amit Deshpande. 2021.
\newblock \href {https://doi.org/10.1609/aaai.v35i14.17505} {How linguistically
  fair are multilingual pre-trained language models?}
\newblock \emph{Proceedings of the AAAI Conference on Artificial Intelligence},
  35(14):12710--12718.

\bibitem[{Conneau and Lample(2019)}]{NEURIPS2019_c04c19c2}
Alexis Conneau and Guillaume Lample. 2019.
\newblock \href
  {https://proceedings.neurips.cc/paper_files/paper/2019/file/c04c19c2c2474dbf5f7ac4372c5b9af1-Paper.pdf}
  {Cross-lingual language model pretraining}.
\newblock In \emph{Advances in Neural Information Processing Systems},
  volume~32. Curran Associates, Inc.

\bibitem[{Conneau et~al.(2018)Conneau, Rinott, Lample, Williams, Bowman,
  Schwenk, and Stoyanov}]{conneau-etal-2018-xnli}
Alexis Conneau, Ruty Rinott, Guillaume Lample, Adina Williams, Samuel Bowman,
  Holger Schwenk, and Veselin Stoyanov. 2018.
\newblock \href {https://doi.org/10.18653/v1/D18-1269} {{XNLI}: Evaluating
  cross-lingual sentence representations}.
\newblock In \emph{Proceedings of the 2018 Conference on Empirical Methods in
  Natural Language Processing}, pages 2475--2485, Brussels, Belgium.
  Association for Computational Linguistics.

\bibitem[{Conneau et~al.(2020)Conneau, Wu, Li, Zettlemoyer, and
  Stoyanov}]{conneau-etal-2020-emerging}
Alexis Conneau, Shijie Wu, Haoran Li, Luke Zettlemoyer, and Veselin Stoyanov.
  2020.
\newblock \href {https://doi.org/10.18653/v1/2020.acl-main.536} {Emerging
  cross-lingual structure in pretrained language models}.
\newblock In \emph{Proceedings of the 58th Annual Meeting of the Association
  for Computational Linguistics}, pages 6022--6034, Online. Association for
  Computational Linguistics.

\bibitem[{Devlin et~al.(2019)Devlin, Chang, Lee, and
  Toutanova}]{devlin-etal-2019-bert}
Jacob Devlin, Ming-Wei Chang, Kenton Lee, and Kristina Toutanova. 2019.
\newblock \href {https://doi.org/10.18653/v1/N19-1423} {{BERT}: Pre-training of
  deep bidirectional transformers for language understanding}.
\newblock In \emph{Proceedings of the 2019 Conference of the North {A}merican
  Chapter of the Association for Computational Linguistics: Human Language
  Technologies, Volume 1 (Long and Short Papers)}, pages 4171--4186,
  Minneapolis, Minnesota. Association for Computational Linguistics.

\bibitem[{ElSahar and El-Beltagy(2015)}]{elsahar2015building}
Hady ElSahar and Samhaa~R El-Beltagy. 2015.
\newblock Building large arabic multi-domain resources for sentiment analysis.
\newblock In \emph{International conference on intelligent text processing and
  computational linguistics}, pages 23--34. Springer.

\bibitem[{Fang et~al.(2021)Fang, Wang, Gan, Sun, and Liu}]{fang2021filter}
Yuwei Fang, Shuohang Wang, Zhe Gan, Siqi Sun, and Jingjing Liu. 2021.
\newblock Filter: An enhanced fusion method for cross-lingual language
  understanding.
\newblock In \emph{Proceedings of the AAAI Conference on Artificial
  Intelligence}, volume~35, pages 12776--12784.

\bibitem[{Fetahu et~al.(2023)Fetahu, Chen, Kar, Rokhlenko, and
  Malmasi}]{fetahu-etal-2023-multiconer}
Besnik Fetahu, Zhiyu Chen, Sudipta Kar, Oleg Rokhlenko, and Shervin Malmasi.
  2023.
\newblock \href {https://doi.org/10.18653/v1/2023.findings-emnlp.134}
  {{M}ulti{C}o{NER} v2: a large multilingual dataset for fine-grained and noisy
  named entity recognition}.
\newblock In \emph{Findings of the Association for Computational Linguistics:
  EMNLP 2023}, pages 2027--2051, Singapore. Association for Computational
  Linguistics.

\bibitem[{Fierro and S{\o}gaard(2022)}]{fierro-sogaard-2022-factual}
Constanza Fierro and Anders S{\o}gaard. 2022.
\newblock \href {https://doi.org/10.18653/v1/2022.findings-acl.240} {Factual
  consistency of multilingual pretrained language models}.
\newblock In \emph{Findings of the Association for Computational Linguistics:
  ACL 2022}, pages 3046--3052, Dublin, Ireland. Association for Computational
  Linguistics.

\bibitem[{Gorman and Bedrick(2019)}]{gorman-bedrick-2019-need}
Kyle Gorman and Steven Bedrick. 2019.
\newblock \href {https://doi.org/10.18653/v1/P19-1267} {We need to talk about
  standard splits}.
\newblock In \emph{Proceedings of the 57th Annual Meeting of the Association
  for Computational Linguistics}, pages 2786--2791, Florence, Italy.
  Association for Computational Linguistics.

\bibitem[{Guillaume et~al.(2019)Guillaume, de~Marneffe, and
  Perrier}]{guillaume-etal-2019-conversion}
Bruno Guillaume, Marie-Catherine de~Marneffe, and Guy Perrier. 2019.
\newblock \href {https://aclanthology.org/2019.tal-2.4} {Conversion et
  am{\'e}liorations de corpus du fran{\c{c}}ais annot{\'e}s en {U}niversal
  {D}ependencies [conversion and improvement of {U}niversal {D}ependencies
  {F}rench corpora]}.
\newblock \emph{Traitement Automatique des Langues}, 60(2):71--95.

\bibitem[{H{\"a}mmerl et~al.(2023)H{\"a}mmerl, Fastowski, Libovick{\'y}, and
  Fraser}]{haemmerl-etal-2023-exploring}
Katharina H{\"a}mmerl, Alina Fastowski, Jind{\v{r}}ich Libovick{\'y}, and
  Alexander Fraser. 2023.
\newblock \href {https://doi.org/10.18653/v1/2023.findings-acl.439} {Exploring
  anisotropy and outliers in multilingual language models for cross-lingual
  semantic sentence similarity}.
\newblock In \emph{Findings of the Association for Computational Linguistics:
  ACL 2023}, pages 7023--7037, Toronto, Canada. Association for Computational
  Linguistics.

\bibitem[{Hou et~al.(2024)Hou, Li, He, Yan, Chen, and
  McAuley}]{hou2024bridging}
Yupeng Hou, Jiacheng Li, Zhankui He, An~Yan, Xiusi Chen, and Julian McAuley.
  2024.
\newblock Bridging language and items for retrieval and recommendation.
\newblock \emph{arXiv preprint arXiv:2403.03952}.

\bibitem[{Ji et~al.(2024)Ji, Mickus, Segonne, and
  Tiedemann}]{ji-etal-2024-machine-translation}
Shaoxiong Ji, Timothee Mickus, Vincent Segonne, and J{\"o}rg Tiedemann. 2024.
\newblock \href {https://aclanthology.org/2024.lrec-main.250} {Can machine
  translation bridge multilingual pretraining and cross-lingual transfer
  learning?}
\newblock In \emph{Proceedings of the 2024 Joint International Conference on
  Computational Linguistics, Language Resources and Evaluation (LREC-COLING
  2024)}, pages 2809--2818, Torino, Italia. ELRA and ICCL.

\bibitem[{Kale et~al.(2021)Kale, Siddhant, Al-Rfou, Xue, Constant, and
  Johnson}]{kale-etal-2021-nmt5}
Mihir Kale, Aditya Siddhant, Rami Al-Rfou, Linting Xue, Noah Constant, and
  Melvin Johnson. 2021.
\newblock \href {https://doi.org/10.18653/v1/2021.acl-short.87} {nm{T}5 - is
  parallel data still relevant for pre-training massively multilingual language
  models?}
\newblock In \emph{Proceedings of the 59th Annual Meeting of the Association
  for Computational Linguistics and the 11th International Joint Conference on
  Natural Language Processing (Volume 2: Short Papers)}, pages 683--691,
  Online. Association for Computational Linguistics.

\bibitem[{Kingma and Ba(2017)}]{kingma2017adam}
Diederik~P. Kingma and Jimmy Ba. 2017.
\newblock \href {http://arxiv.org/abs/1412.6980} {Adam: A method for stochastic
  optimization}.

\bibitem[{Lee et~al.(2017)Lee, Leung, and Li}]{lee-etal-2017-towards}
John Lee, Herman Leung, and Keying Li. 2017.
\newblock \href {https://aclanthology.org/W17-0408} {Towards {U}niversal
  {D}ependencies for learner {C}hinese}.
\newblock In \emph{Proceedings of the {N}o{D}a{L}i{D}a 2017 Workshop on
  Universal Dependencies ({UDW} 2017)}, pages 67--71, Gothenburg, Sweden.
  Association for Computational Linguistics.

\bibitem[{Lewis et~al.(2020)Lewis, Liu, Goyal, Ghazvininejad, Mohamed, Levy,
  Stoyanov, and Zettlemoyer}]{lewis-etal-2020-bart}
Mike Lewis, Yinhan Liu, Naman Goyal, Marjan Ghazvininejad, Abdelrahman Mohamed,
  Omer Levy, Veselin Stoyanov, and Luke Zettlemoyer. 2020.
\newblock \href {https://doi.org/10.18653/v1/2020.acl-main.703} {{BART}:
  Denoising sequence-to-sequence pre-training for natural language generation,
  translation, and comprehension}.
\newblock In \emph{Proceedings of the 58th Annual Meeting of the Association
  for Computational Linguistics}, pages 7871--7880, Online. Association for
  Computational Linguistics.

\bibitem[{Li et~al.(2022)Li, Kim, Bruno, and Zeman}]{UDChinesePatentChar}
Yixuan Li, Gerdes Kim, Guillaume Bruno, and Dan Zeman. 2022.
\newblock \href
  {https://github.com/UniversalDependencies/UD_Chinese-PatentChar} {Ud chinese
  patentchar}.

\bibitem[{Liu et~al.(2020)Liu, Gu, Goyal, Li, Edunov, Ghazvininejad, Lewis, and
  Zettlemoyer}]{liu-etal-2020-multilingual-denoising}
Yinhan Liu, Jiatao Gu, Naman Goyal, Xian Li, Sergey Edunov, Marjan
  Ghazvininejad, Mike Lewis, and Luke Zettlemoyer. 2020.
\newblock \href {https://doi.org/10.1162/tacl_a_00343} {Multilingual denoising
  pre-training for neural machine translation}.
\newblock \emph{Transactions of the Association for Computational Linguistics},
  8:726--742.

\bibitem[{Loshchilov and Hutter(2017)}]{loshchilov2017decoupled}
Ilya Loshchilov and Frank Hutter. 2017.
\newblock Decoupled weight decay regularization.
\newblock \emph{arXiv preprint arXiv:1711.05101}.

\bibitem[{Lyashevskaya et~al.(2018)Lyashevskaya, Rudina, Vlasova, and
  Zhuravleva}]{UDRussianTaiga}
Olga Lyashevskaya, Olga Rudina, Natalia Vlasova, and Anna Zhuravleva. 2018.
\newblock \href {https://github.com/UniversalDependencies/UD_Russian-Taiga} {Ud
  russian taiga}.

\bibitem[{Malmasi et~al.(2022)Malmasi, Fang, Fetahu, Kar, and
  Rokhlenko}]{malmasi-etal-2022-multiconer}
Shervin Malmasi, Anjie Fang, Besnik Fetahu, Sudipta Kar, and Oleg Rokhlenko.
  2022.
\newblock \href {https://aclanthology.org/2022.coling-1.334} {{M}ulti{C}o{NER}:
  A large-scale multilingual dataset for complex named entity recognition}.
\newblock In \emph{Proceedings of the 29th International Conference on
  Computational Linguistics}, pages 3798--3809, Gyeongju, Republic of Korea.
  International Committee on Computational Linguistics.

\bibitem[{McDonald et~al.(2013)McDonald, Nivre, Quirmbach-Brundage, Goldberg,
  Das, Ganchev, Hall, Petrov, Zhang, T{\"a}ckstr{\"o}m, Bedini,
  Bertomeu~Castell{\'o}, and Lee}]{mcdonald-etal-2013-universal}
Ryan McDonald, Joakim Nivre, Yvonne Quirmbach-Brundage, Yoav Goldberg, Dipanjan
  Das, Kuzman Ganchev, Keith Hall, Slav Petrov, Hao Zhang, Oscar
  T{\"a}ckstr{\"o}m, Claudia Bedini, N{\'u}ria Bertomeu~Castell{\'o}, and
  Jungmee Lee. 2013.
\newblock \href {https://aclanthology.org/P13-2017} {{U}niversal {D}ependency
  annotation for multilingual parsing}.
\newblock In \emph{Proceedings of the 51st Annual Meeting of the Association
  for Computational Linguistics (Volume 2: Short Papers)}, pages 92--97, Sofia,
  Bulgaria. Association for Computational Linguistics.

\bibitem[{Mohit et~al.(2012{\natexlab{a}})Mohit, Schneider, Bhowmick, Oflazer,
  and Smith}]{mohit-etal-2012-recall}
Behrang Mohit, Nathan Schneider, Rishav Bhowmick, Kemal Oflazer, and Noah~A.
  Smith. 2012{\natexlab{a}}.
\newblock \href {https://aclanthology.org/E12-1017} {Recall-oriented learning
  of named entities in {A}rabic {W}ikipedia}.
\newblock In \emph{Proceedings of the 13th Conference of the {E}uropean Chapter
  of the Association for Computational Linguistics}, pages 162--173, Avignon,
  France. Association for Computational Linguistics.

\bibitem[{Mohit et~al.(2012{\natexlab{b}})Mohit, Schneider, Bhowmick, Oflazer,
  and Smith}]{mohit2012recall}
Behrang Mohit, Nathan Schneider, Rishav Bhowmick, Kemal Oflazer, and Noah~A
  Smith. 2012{\natexlab{b}}.
\newblock Recall-oriented learning of named entities in arabic wikipedia.
\newblock In \emph{Proceedings of the 13th Conference of the European Chapter
  of the Association for Computational Linguistics}, pages 162--173.

\bibitem[{Muennighoff et~al.(2022)Muennighoff, Wang, Sutawika, Roberts,
  Biderman, Scao, Bari, Shen, Yong, Schoelkopf
  et~al.}]{muennighoff2022crosslingual}
Niklas Muennighoff, Thomas Wang, Lintang Sutawika, Adam Roberts, Stella
  Biderman, Teven~Le Scao, M~Saiful Bari, Sheng Shen, Zheng-Xin Yong, Hailey
  Schoelkopf, et~al. 2022.
\newblock Crosslingual generalization through multitask finetuning.
\newblock \emph{arXiv preprint arXiv:2211.01786}.

\bibitem[{Nivre et~al.(2017)Nivre, Agi{\'c}, Ahrenberg, Antonsen, Aranzabe,
  Asahara, Ateyah, Attia, Atutxa, Badmaeva, Ballesteros, Banerjee, Bank, Bauer,
  Bengoetxea, Bhat, Bick, Bosco, Bouma, Bowman, Burchardt, Candito, Caron,
  Cebiro{\u g}lu~Eryi{\u g}it, Celano, Cetin, Chalub, Choi, Cho, Cinkov{\'a},
  {\c C}{\"o}ltekin, Connor, de~Marneffe, de~Paiva, Diaz~de Ilarraza,
  Dobrovoljc, Dozat, Droganova, Eli, Elkahky, Erjavec, Farkas,
  Fernandez~Alcalde, Foster, Freitas, Gajdo{\v s}ov{\'a}, Galbraith, Garcia,
  Ginter, Goenaga, Gojenola, G{\"o}k{\i}rmak, Goldberg, G{\'o}mez~Guinovart,
  Gonz{\'a}les~Saavedra, Grioni, Gr{\= u}z{\={\i}}tis, Guillaume, Habash,
  Haji{\v c}, Haji{\v c}~jr., H{\`a}~M{\~y}, Harris, Haug, Hladk{\'a},
  Hlav{\'a}{\v c}ov{\'a}, Hohle, Ion, Irimia, Johannsen, J{\o}rgensen, Ka{\c
  s}{\i}kara, Kanayama, Kanerva, Kayadelen, Kettnerov{\'a}, Kirchner, Kotsyba,
  Krek, Kwak, Laippala, Lambertino, Lando, L{\^e}~H{\`{\^o}}ng, Lenci,
  Lertpradit, Leung, Li, Li, Ljube{\v s}i{\'c}, Loginova, Lyashevskaya, Lynn,
  Macketanz, Makazhanov, Mandl, Manning, Manurung, M{\u a}r{\u a}nduc, Mare{\v
  c}ek, Marheinecke, Mart{\'{\i}}nez~Alonso, Martins, Ma{\v s}ek, Matsumoto,
  {McDonald}, Mendon{\c c}a, Missil{\"a}, Mititelu, Miyao, Montemagni, More,
  Moreno~Romero, Mori, Moskalevskyi, Muischnek, Mustafina, M{\"u}{\"u}risep,
  Nainwani, Nedoluzhko, Nguy{\~{\^e}}n~Th{\d i}, Nguy{\~{\^e}}n Th{\d i}~Minh,
  Nikolaev, Nitisaroj, Nurmi, Ojala, Osenova, {\O}vrelid, Pascual, Passarotti,
  Perez, Perrier, Petrov, Piitulainen, Pitler, Plank, Popel, Pretkalni{\c n}a,
  Prokopidis, Puolakainen, Pyysalo, Rademaker, Real, Reddy, Rehm, Rinaldi,
  Rituma, Rosa, Rovati, Saleh, Sanguinetti, Saul{\={\i}}te, Sawanakunanon,
  Schuster, Seddah, Seeker, Seraji, Shakurova, Shen, Shimada, Shohibussirri,
  Silveira, Simi, Simionescu, Simk{\'o}, {\v S}imkov{\'a}, Simov, Smith,
  Stella, Strnadov{\'a}, Suhr, Sulubacak, Sz{\'a}nt{\'o}, Taji, Tanaka,
  Trosterud, Trukhina, Tsarfaty, Tyers, Uematsu, Ure{\v s}ov{\'a}, Uria,
  Uszkoreit, van Noord, Varga, Vincze, Washington, Yu, {\v Z}abokrtsk{\'y},
  Zeman, and Zhu}]{11234/1-2184}
Joakim Nivre, {\v Z}eljko Agi{\'c}, Lars Ahrenberg, Lene Antonsen, Maria~Jesus
  Aranzabe, Masayuki Asahara, Luma Ateyah, Mohammed Attia, Aitziber Atutxa,
  Elena Badmaeva, Miguel Ballesteros, Esha Banerjee, Sebastian Bank, John
  Bauer, Kepa Bengoetxea, Riyaz~Ahmad Bhat, Eckhard Bick, Cristina Bosco, Gosse
  Bouma, Sam Bowman, Aljoscha Burchardt, Marie Candito, Gauthier Caron,
  G{\"u}l{\c s}en Cebiro{\u g}lu~Eryi{\u g}it, Giuseppe G.~A. Celano, Savas
  Cetin, Fabricio Chalub, Jinho Choi, Yongseok Cho, Silvie Cinkov{\'a}, {\c
  C}a{\u g}r{\i} {\c C}{\"o}ltekin, Miriam Connor, Marie-Catherine de~Marneffe,
  Valeria de~Paiva, Arantza Diaz~de Ilarraza, Kaja Dobrovoljc, Timothy Dozat,
  Kira Droganova, Marhaba Eli, Ali Elkahky, Toma{\v z} Erjavec, Rich{\'a}rd
  Farkas, Hector Fernandez~Alcalde, Jennifer Foster, Cl{\'a}udia Freitas,
  Katar{\'{\i}}na Gajdo{\v s}ov{\'a}, Daniel Galbraith, Marcos Garcia, Filip
  Ginter, Iakes Goenaga, Koldo Gojenola, Memduh G{\"o}k{\i}rmak, Yoav Goldberg,
  Xavier G{\'o}mez~Guinovart, Berta Gonz{\'a}les~Saavedra, Matias Grioni,
  Normunds Gr{\= u}z{\={\i}}tis, Bruno Guillaume, Nizar Habash, Jan Haji{\v c},
  Jan Haji{\v c}~jr., Linh H{\`a}~M{\~y}, Kim Harris, Dag Haug, Barbora
  Hladk{\'a}, Jaroslava Hlav{\'a}{\v c}ov{\'a}, Petter Hohle, Radu Ion, Elena
  Irimia, Anders Johannsen, Fredrik J{\o}rgensen, H{\"u}ner Ka{\c s}{\i}kara,
  Hiroshi Kanayama, Jenna Kanerva, Tolga Kayadelen, V{\'a}clava Kettnerov{\'a},
  Jesse Kirchner, Natalia Kotsyba, Simon Krek, Sookyoung Kwak, Veronika
  Laippala, Lorenzo Lambertino, Tatiana Lando, Phương L{\^e}~H{\`{\^o}}ng,
  Alessandro Lenci, Saran Lertpradit, Herman Leung, Cheuk~Ying Li, Josie Li,
  Nikola Ljube{\v s}i{\'c}, Olga Loginova, Olga Lyashevskaya, Teresa Lynn,
  Vivien Macketanz, Aibek Makazhanov, Michael Mandl, Christopher Manning, Ruli
  Manurung, C{\u a}t{\u a}lina M{\u a}r{\u a}nduc, David Mare{\v c}ek, Katrin
  Marheinecke, H{\'e}ctor Mart{\'{\i}}nez~Alonso, Andr{\'e} Martins, Jan Ma{\v
  s}ek, Yuji Matsumoto, Ryan {McDonald}, Gustavo Mendon{\c c}a, Anna
  Missil{\"a}, Verginica Mititelu, Yusuke Miyao, Simonetta Montemagni, Amir
  More, Laura Moreno~Romero, Shunsuke Mori, Bohdan Moskalevskyi, Kadri
  Muischnek, Nina Mustafina, Kaili M{\"u}{\"u}risep, Pinkey Nainwani, Anna
  Nedoluzhko, Lương Nguy{\~{\^e}}n~Th{\d i}, Huy{\`{\^e}}n Nguy{\~{\^e}}n
  Th{\d i}~Minh, Vitaly Nikolaev, Rattima Nitisaroj, Hanna Nurmi, Stina Ojala,
  Petya Osenova, Lilja {\O}vrelid, Elena Pascual, Marco Passarotti,
  Cenel-Augusto Perez, Guy Perrier, Slav Petrov, Jussi Piitulainen, Emily
  Pitler, Barbara Plank, Martin Popel, Lauma Pretkalni{\c n}a, Prokopis
  Prokopidis, Tiina Puolakainen, Sampo Pyysalo, Alexandre Rademaker, Livy Real,
  Siva Reddy, Georg Rehm, Larissa Rinaldi, Laura Rituma, Rudolf Rosa, Davide
  Rovati, Shadi Saleh, Manuela Sanguinetti, Baiba Saul{\={\i}}te, Yanin
  Sawanakunanon, Sebastian Schuster, Djam{\'e} Seddah, Wolfgang Seeker, Mojgan
  Seraji, Lena Shakurova, Mo~Shen, Atsuko Shimada, Muh Shohibussirri, Natalia
  Silveira, Maria Simi, Radu Simionescu, Katalin Simk{\'o}, M{\'a}ria {\v
  S}imkov{\'a}, Kiril Simov, Aaron Smith, Antonio Stella, Jana Strnadov{\'a},
  Alane Suhr, Umut Sulubacak, Zsolt Sz{\'a}nt{\'o}, Dima Taji, Takaaki Tanaka,
  Trond Trosterud, Anna Trukhina, Reut Tsarfaty, Francis Tyers, Sumire Uematsu,
  Zde{\v n}ka Ure{\v s}ov{\'a}, Larraitz Uria, Hans Uszkoreit, Gertjan van
  Noord, Viktor Varga, Veronika Vincze, Jonathan~North Washington, Zhuoran Yu,
  Zden{\v e}k {\v Z}abokrtsk{\'y}, Daniel Zeman, and Hanzhi Zhu. 2017.
\newblock \href {http://hdl.handle.net/11234/1-2184} {Universal dependencies
  2.0 – {CoNLL} 2017 shared task development and test data}.
\newblock {LINDAT}/{CLARIAH}-{CZ} digital library at the Institute of Formal
  and Applied Linguistics ({{\'U}FAL}), Faculty of Mathematics and Physics,
  Charles University.

\bibitem[{Nivre et~al.(2020)Nivre, De~Marneffe, Ginter, Haji{\v{c}}, Manning,
  Pyysalo, Schuster, Tyers, and Zeman}]{nivre2020universal}
Joakim Nivre, Marie-Catherine De~Marneffe, Filip Ginter, Jan Haji{\v{c}},
  Christopher~D Manning, Sampo Pyysalo, Sebastian Schuster, Francis Tyers, and
  Daniel Zeman. 2020.
\newblock Universal dependencies v2: An evergrowing multilingual treebank
  collection.
\newblock \emph{arXiv preprint arXiv:2004.10643}.

\bibitem[{Ott et~al.(2019)Ott, Edunov, Baevski, Fan, Gross, Ng, Grangier, and
  Auli}]{ott-etal-2019-fairseq}
Myle Ott, Sergey Edunov, Alexei Baevski, Angela Fan, Sam Gross, Nathan Ng,
  David Grangier, and Michael Auli. 2019.
\newblock \href {https://doi.org/10.18653/v1/N19-4009} {fairseq: A fast,
  extensible toolkit for sequence modeling}.
\newblock In \emph{Proceedings of the 2019 Conference of the North {A}merican
  Chapter of the Association for Computational Linguistics (Demonstrations)},
  pages 48--53, Minneapolis, Minnesota. Association for Computational
  Linguistics.

\bibitem[{Qi et~al.(2019)Qi, Yasuoka, and Zeman}]{UDChineseGSDSimp}
Peng Qi, Koichi Yasuoka, and Dan Zeman. 2019.
\newblock \href {https://github.com/UniversalDependencies/UD_Chinese-GSDSimp}
  {Ud chinese gsdsimp}.

\bibitem[{Radford et~al.(2019)Radford, Wu, Child, Luan, Amodei, Sutskever
  et~al.}]{radford2019language}
Alec Radford, Jeffrey Wu, Rewon Child, David Luan, Dario Amodei, Ilya
  Sutskever, et~al. 2019.
\newblock Language models are unsupervised multitask learners.
\newblock \emph{OpenAI blog}, 1(8):9.

\bibitem[{Rust et~al.(2021)Rust, Pfeiffer, Vuli{\'c}, Ruder, and
  Gurevych}]{rust-etal-2021-good}
Phillip Rust, Jonas Pfeiffer, Ivan Vuli{\'c}, Sebastian Ruder, and Iryna
  Gurevych. 2021.
\newblock \href {https://doi.org/10.18653/v1/2021.acl-long.243} {How good is
  your tokenizer? on the monolingual performance of multilingual language
  models}.
\newblock In \emph{Proceedings of the 59th Annual Meeting of the Association
  for Computational Linguistics and the 11th International Joint Conference on
  Natural Language Processing (Volume 1: Long Papers)}, pages 3118--3135,
  Online. Association for Computational Linguistics.

\bibitem[{Sennrich et~al.(2016)Sennrich, Haddow, and
  Birch}]{sennrich-etal-2016-neural}
Rico Sennrich, Barry Haddow, and Alexandra Birch. 2016.
\newblock \href {https://doi.org/10.18653/v1/P16-1162} {Neural machine
  translation of rare words with subword units}.
\newblock In \emph{Proceedings of the 54th Annual Meeting of the Association
  for Computational Linguistics (Volume 1: Long Papers)}, pages 1715--1725,
  Berlin, Germany. Association for Computational Linguistics.

\bibitem[{Siddhant et~al.(2020)Siddhant, Johnson, Tsai, Ari, Riesa, Bapna,
  Firat, and Raman}]{siddhant2020evaluating}
Aditya Siddhant, Melvin Johnson, Henry Tsai, Naveen Ari, Jason Riesa, Ankur
  Bapna, Orhan Firat, and Karthik Raman. 2020.
\newblock Evaluating the cross-lingual effectiveness of massively multilingual
  neural machine translation.
\newblock In \emph{Proceedings of the AAAI conference on artificial
  intelligence}, volume~34, pages 8854--8861.

\bibitem[{Smetanin and Komarov(2019)}]{Smetanin-SA-2019}
Sergey Smetanin and Michail Komarov. 2019.
\newblock \href {https://doi.org/10.1109/CBI.2019.00062} {Sentiment analysis of
  product reviews in russian using convolutional neural networks}.
\newblock In \emph{2019 IEEE 21st Conference on Business Informatics (CBI)},
  volume~01, pages 482--486.

\bibitem[{Tiedemann(2012)}]{tiedemann2012parallel}
J{\"o}rg Tiedemann. 2012.
\newblock Parallel data, tools and interfaces in opus.
\newblock In \emph{Proceedings of LREC}, volume 2012, pages 2214--2218.

\bibitem[{Vaswani et~al.(2017)Vaswani, Shazeer, Parmar, Uszkoreit, Jones,
  Gomez, Kaiser, and Polosukhin}]{NIPS2017_3f5ee243}
Ashish Vaswani, Noam Shazeer, Niki Parmar, Jakob Uszkoreit, Llion Jones,
  Aidan~N Gomez, \L~ukasz Kaiser, and Illia Polosukhin. 2017.
\newblock \href
  {https://proceedings.neurips.cc/paper_files/paper/2017/file/3f5ee243547dee91fbd053c1c4a845aa-Paper.pdf}
  {Attention is all you need}.
\newblock In \emph{Advances in Neural Information Processing Systems},
  volume~30. Curran Associates, Inc.

\bibitem[{Wang et~al.(2019)Wang, Che, Guo, Liu, and Liu}]{wang2019cross}
Yuxuan Wang, Wanxiang Che, Jiang Guo, Yijia Liu, and Ting Liu. 2019.
\newblock Cross-lingual bert transformation for zero-shot dependency parsing.
\newblock In \emph{Proceedings of the 2019 Conference on Empirical Methods in
  Natural Language Processing and the 9th International Joint Conference on
  Natural Language Processing (EMNLP-IJCNLP)}, pages 5721--5727.

\bibitem[{Williams et~al.(2018)Williams, Nangia, and
  Bowman}]{williams-etal-2018-broad}
Adina Williams, Nikita Nangia, and Samuel Bowman. 2018.
\newblock \href {https://doi.org/10.18653/v1/N18-1101} {A broad-coverage
  challenge corpus for sentence understanding through inference}.
\newblock In \emph{Proceedings of the 2018 Conference of the North {A}merican
  Chapter of the Association for Computational Linguistics: Human Language
  Technologies, Volume 1 (Long Papers)}, pages 1112--1122, New Orleans,
  Louisiana. Association for Computational Linguistics.

\bibitem[{Wong et~al.(2017)Wong, Gerdes, Leung, and
  Lee}]{wong-etal-2017-quantitative}
Tak-sum Wong, Kim Gerdes, Herman Leung, and John Lee. 2017.
\newblock \href {https://aclanthology.org/W17-6530} {Quantitative comparative
  syntax on the {C}antonese-{M}andarin parallel dependency treebank}.
\newblock In \emph{Proceedings of the Fourth International Conference on
  Dependency Linguistics (Depling 2017)}, pages 266--275, Pisa, Italy.
  Link{\"o}ping University Electronic Press.

\bibitem[{Wu and Dredze(2019)}]{wu2019beto}
Shijie Wu and Mark Dredze. 2019.
\newblock Beto, bentz, becas: The surprising cross-lingual effectiveness of
  bert.
\newblock In \emph{Proceedings of the 2019 Conference on Empirical Methods in
  Natural Language Processing and the 9th International Joint Conference on
  Natural Language Processing (EMNLP-IJCNLP)}, pages 833--844.

\bibitem[{Wu and Dredze(2020)}]{wu2020explicit}
Shijie Wu and Mark Dredze. 2020.
\newblock Do explicit alignments robustly improve multilingual encoders?
\newblock In \emph{Proceedings of the 2020 Conference on Empirical Methods in
  Natural Language Processing (EMNLP)}, pages 4471--4482.

\bibitem[{Xue et~al.(2021)Xue, Constant, Roberts, Kale, Al-Rfou, Siddhant,
  Barua, and Raffel}]{xue-etal-2021-mt5}
Linting Xue, Noah Constant, Adam Roberts, Mihir Kale, Rami Al-Rfou, Aditya
  Siddhant, Aditya Barua, and Colin Raffel. 2021.
\newblock \href {https://doi.org/10.18653/v1/2021.naacl-main.41} {m{T}5: A
  massively multilingual pre-trained text-to-text transformer}.
\newblock In \emph{Proceedings of the 2021 Conference of the North American
  Chapter of the Association for Computational Linguistics: Human Language
  Technologies}, pages 483--498, Online. Association for Computational
  Linguistics.

\bibitem[{Zeldes(2017)}]{Zeldes2017}
Amir Zeldes. 2017.
\newblock \href {https://doi.org/http://dx.doi.org/10.1007/s10579-016-9343-x}
  {The {GUM} corpus: Creating multilayer resources in the classroom}.
\newblock \emph{Language Resources and Evaluation}, 51(3):581--612.

\bibitem[{Zeman et~al.(2023)Zeman, Guiller, and Guillaume}]{UDChineseBeginner}
Dan Zeman, Kirian Guiller, and Bruno Guillaume. 2023.
\newblock \href {https://github.com/UniversalDependencies/UD_Chinese-Beginner}
  {Ud chinese beginner}.

\bibitem[{Zem{\' a}nek(2008)}]{Zemnek2008DependencyT}
Otakar Smr{\v z} Viktor Bielick{\' y} Iveta Kouřilov{\' a} Jakub~Kr{\' a}{\v
  c}mar Zem{\' a}nek. 2008.
\newblock \href {https://api.semanticscholar.org/CorpusID:2374836} {Dependency
  treebank : A word on the million words}.

\bibitem[{Ziemski et~al.(2016)Ziemski, Junczys-Dowmunt, and
  Pouliquen}]{ziemski-etal-2016-united}
Micha{\l} Ziemski, Marcin Junczys-Dowmunt, and Bruno Pouliquen. 2016.
\newblock \href {https://aclanthology.org/L16-1561} {The {U}nited {N}ations
  parallel corpus v1.0}.
\newblock In \emph{Proceedings of the Tenth International Conference on
  Language Resources and Evaluation ({LREC}'16)}, pages 3530--3534,
  Portoro{\v{z}}, Slovenia. European Language Resources Association (ELRA).

\end{thebibliography}
\bibliographystyle{acl_natbib}

\clearpage
\appendix

\section{Overview of pretraining objectives}
\label{adx:objectives}

\begin{table*}[!htbp]
    \scriptsize
    \begin{tabular}{p{0.15\linewidth} p{0.375\linewidth} p{0.375\linewidth}}
    \toprule
        Objective & \textbf{Source input} & \textbf{Target output} \\
    \midrule
        {{\vfill\textbf{2-LM}\vfill}} & \tt ▁D'autres ▁mesures ▁de ▁ce ▁type ▁vont ▁être [MASK] [MASK], ▁en ▁coopération ▁avec ▁d'autres ▁associations ▁de ▁Rom s, ▁de ▁Sin tis ▁et ▁de [MASK] ▁du ▁voyage ▁(<< C am min anti >> ).  </s> &	\tt <s> ▁D'autres ▁mesures ▁de ▁ce ▁type ▁vont ▁être ▁appliqu ées, ▁en ▁coopération ▁avec ▁d'autres ▁associations ▁de ▁Rom s, ▁de ▁Sin tis ▁et ▁de ▁gens ▁du ▁voyage ▁(<< C am min anti >> ). </s> \\
    \rowcolor{gray!20}
        {{\vfill\textbf{2-MT}\vfill}} & \tt <fr> ▁D'autres ▁mesures ▁de ▁ce ▁type ▁vont ▁être ▁appliqu ées, ▁en ▁coopération ▁avec ▁d'autres ▁associations ▁de ▁Rom s, ▁de ▁Sin tis ▁et ▁de ▁gens ▁du ▁voyage ▁(<< C am min anti >> ). & \tt <s> ▁Other ▁similar ▁measures ▁are ▁going ▁to ▁be ▁taken ▁in ▁cooperation ▁with ▁other ▁Rom a, ▁Sin ti ▁and ▁Travel lers ▁(" C am min anti ") ▁associ ations. </s> \\ 
    \midrule
        {{\vfill\textbf{CLM}\vfill}} &  \tt  ...  ▁Divers ▁accords ▁ad ▁hoc ▁ont ▁été ▁conclus ▁à ▁cet ▁effet ▁par ▁le ▁Ministère ▁de ▁l'éducation ▁et ▁l'as sociation ▁Op era ▁Nom ad i. ▁D'autres ▁mesures ▁de ▁ce ▁type ▁vont ▁être ▁appliqu ées, ▁en ▁coopération ▁avec ▁d'autres ▁associations ▁de ▁Rom s, ▁de ▁Sin tis ▁et ▁de ▁gens ▁du ▁voyage ▁(<< C am min anti >> ).  ... & \tt ... ▁accords ▁ad ▁hoc ▁ont ▁été ▁conclus ▁à ▁cet ▁effet ▁par ▁le ▁Ministère ▁de ▁l'éducation ▁et ▁l'as sociation ▁Op era ▁Nom ad i. ▁D'autres  ▁mesures ▁de ▁ce ▁type ▁vont ▁être ▁appliqu ées, ▁en ▁coopération ▁avec ▁d'autres ▁associations ▁de ▁Rom s, ▁de ▁Sin tis ▁et ▁de ▁gens ▁du ▁voyage ▁(<< C am min anti >> ).  ...  \\
    \rowcolor{gray!20}
        {{\vfill\textbf{TLM}\vfill}} & \tt ▁D'autres ▁mesures ▁de ▁ce ▁type ▁vont ▁être ▁appliqu ées, ▁en ▁coopération ▁avec ▁d'autres ▁associations ▁de ▁Rom s, ▁de ▁Sin tis ▁et ▁de ▁gens ▁du ▁voyage ▁(<< C am min anti >> ). <fr2en> ▁Other ▁similar ▁measures ▁are ▁going ▁to ▁be ▁taken ▁in ▁cooperation ▁with ▁other ▁Rom a, ▁Sin ti ▁and ▁Travel lers ▁(" C am min anti ") ▁associ ations. & \tt ▁mesures ▁de ▁ce ▁type ▁vont ▁être ▁appliqu ées, ▁en ▁coopération ▁avec ▁d'autres ▁associations ▁de ▁Rom s, ▁de ▁Sin tis ▁et ▁de ▁gens ▁du ▁voyage ▁(<< C am min anti >> ). <fr2en> ▁Other ▁similar ▁measures ▁are ▁going ▁to ▁be ▁taken ▁in ▁cooperation ▁with ▁other ▁Rom a, ▁Sin ti ▁and ▁Travel lers ▁(" C am min anti ") ▁associ ations. </s>  \\
        {{\vfill\textbf{MLM}\vfill}} & \tt <s> ▁D'autres ▁mesures ▁de ▁ce ▁type ▁vont ▁être [MASK] [MASK], ▁en ▁coopération ▁avec ▁d'autres ▁associations ▁de ▁Rom s, ▁de ▁Sin tis ▁et ▁de [MASK] ▁du ▁voyage ▁(<< C am min anti >> ).  </s> &	\tt <s> ▁D'autres ▁mesures ▁de ▁ce ▁type ▁vont ▁être ▁appliqu ées, ▁en ▁coopération ▁avec ▁d'autres ▁associations ▁de ▁Rom s, ▁de ▁Sin tis ▁et ▁de ▁gens ▁du ▁voyage ▁(<< C am min anti >> ). </s> \\
    \bottomrule
    \end{tabular}
    \caption{Overview of the different objectives considered in this study. Top two rows: two-stacks (encoder-decoder) models; bottom three rows: single-stack (encoder-only or decoder-only) models.}
    \label{tab:objectives}
\end{table*}

\Cref{tab:objectives} displays an example data point for all pretraining objectives we consider.
In principle, the CLM is a document-level objective, i.e., the full document would be used as an input rather than the two sentences we show here.

\section{Datasets statistics}
\label{adx:downstream-dataset-details}

\begin{table*}[!htbp]
    \centering
    \resizebox{\columnwidth}{!}{
    \begin{tabular}{l S[table-format=9.0] S[table-format=6.0] S[table-format=5.0] S[table-format=9.0] }
        \toprule
        & \textbf{Train} & \textbf{Validation} & \textbf{Test} & \textbf{Total} \\
        \midrule
        \textbf{UNPC}          & 114376177 & 76303 & 40712 & 114493192 \\
        \textbf{OpSub} & 81622353 & 359035 & 77342 & 82058730 \\
        \midrule
        \textbf{Total} & 195998530 & 435338 & 118054 & 196551922 \\
        \bottomrule
    \end{tabular}
    }
    \caption{Number of sentences in pretraining corpora.}
    \label{tab:pretraining-dataset}
\end{table*}

An overview of the volume of data available for pretraining is displayed in \Cref{tab:pretraining-dataset}.
The majority of the data were used for training.

\begin{table*}[!htbp]
\centering
\scriptsize
\begin{adjustbox}{max width=\textwidth}
\begin{tabular}{lllccccc}
\toprule
\multicolumn{1}{c}{\textbf{Task}} & \multicolumn{1}{c}{\textbf{Language}} & \multicolumn{1}{c}{\textbf{Dataset}} & \multicolumn{1}{c}{\textbf{Class Count}} & \multicolumn{1}{c}{\textbf{Train}} & \multicolumn{1}{c}{\textbf{Validation}} & \multicolumn{1}{c}{\textbf{Test}} & \multicolumn{1}{c}{\textbf{Total}} \\
\midrule
\multirow{6}{*}{\textbf{SA}}
& EN & \multirow{4}{*}{Amazon Review~\citep{hou2024bridging}} & 5 & 200000 & 5000 & 5000 & 210000 \\
& ES &  & 5 & 200000 & 5000 & 5000 & 210000 \\
& FR &  & 5 & 200000 & 5000 & 5000 & 210000 \\
& ZH &  & 5 & 200000 & 5000 & 5000 & 210000 \\
& RU & RuReviews~\citep{Smetanin-SA-2019} & 3 & 85601 & 2143 & 2137 & 89881 \\
& AR & \texttt{ar\_res\_reviews}~\citep{elsahar2015building} & 2 & 6680 & 835 & 835 & 8350 \\
\midrule
\multirow{6}{*}{\textbf{NER}} 
& EN & MultiCoNER v2~\citep{fetahu-etal-2023-multiconer} & 3 & 253011 & 13323 & 3773671 & 4040005\\
& ES & MultiCoNER v2 & 3 & 262814 & 13462 & 3925900 & 4202176\\
& FR & MultiCoNER v2 & 3 & 247743 & 13062 & 3742924 & 4003729\\
& ZH & MultiCoNER v2 & 3 & 245606 & 12816 & 489605 & 748027\\
& RU & MultiCoNER v1~\citep{malmasi-etal-2022-multiconer} & 3 & 242384 & 12787 & 2061318 & 2316489\\
& AR & AQMAR Wikipedia NER corpus~\citep{mohit2012recall} & 3 & 57053 & 8615 & 8185 & 73853\\
\midrule
\multirow{11}{*}{\textbf{POS}} 
& EN & UD\_English-GUM~\citep{Zeldes2017} & 16 & 128391 & 16070 & 15554 & 160015\\
& ES & UD\_Spanish-GSD~\citep{mcdonald-etal-2013-universal} & 16 & 127459 & 16916 & 15645 & 160020 \\
& FR & UD\_French-GSD~\citep{guillaume-etal-2019-conversion} & 15 & 127638 & 16207 & 16167 & 160012 \\
& \multirow{6}{*}{ZH}
     & UD\_Chinese-Beginner~\citep{UDChineseBeginner,ChineseGrammarWiki}+ & \multirow{6}{*}{16} & \multirow{6}{*}{128935} & \multirow{6}{*}{15680} & \multirow{6}{*}{15758} & \multirow{6}{*}{160373} \\ 
&    & UD\_Chinese-PUD~\citep{11234/1-2184}+ \\ 
&    & UD\_Chinese-HK~\citep{wong-etal-2017-quantitative}+ \\ 
&    & UD\_Chinese-CFL~\citep{lee-etal-2017-towards}+\\ 
&    & UD\_Chinese-PatentChar~\citep{UDChinesePatentChar}+ \\ 
&    & UD\_Chinese-GSDSmp~\citep{UDChineseGSDSimp} \\
& RU & UD\_Russian-Taiga~\citep{UDRussianTaiga} & 16 & 127647 & 16175 & 16184 & 160006\\
& AR & UD\_Arabic-PADT~\citep{Zemnek2008DependencyT} & 16 & 127552 & 16608 & 15848 & 160008 \\
\midrule
\multirow{6}{*}{\textbf{NLI}}
& EN & \multirow{6}{*}{XNLI~\citep{conneau-etal-2018-xnli}} & 3 & 392702 & 2490 & 5010 & 400202 \\
& ES &  & 3 & 392702 & 2490 & 5010 & 400202 \\
& FR &  & 3 & 392702 & 2490 & 5010 & 400202 \\
& ZH &  & 3 & 392702 & 2490 & 5010 & 400202 \\
& RU &  & 3 & 392702 & 2490 & 5010 & 400202 \\
& AR &  & 3 & 392702 & 2490 & 5010 & 400202 \\
\bottomrule
\end{tabular}
\end{adjustbox}
\caption{Statistics of datasets used for downstream evaluation tasks.}
\label{tab:downstream-dataset-statistics}
\end{table*}

In \Cref{tab:downstream-dataset-statistics}, we present an overview of the datasets used for downstream evaluation.

\section{Detailed results}
\label{adx:benchmarks}

\begin{table*}[!htbp]
    \centering
    \scriptsize
    
	\begin{tabular}{l@{{~~}}l c@{{\smaller$\pm$}}>{\smaller}c c@{{\smaller$\pm$}}>{\smaller}c c@{{\smaller$\pm$}}>{\smaller}c c@{{\smaller$\pm$}}>{\smaller}c c@{{\smaller$\pm$}}>{\smaller}c c@{{\smaller$\pm$}}>{\smaller}c}
    \toprule
    \multicolumn{1}{c}{\multirow{2}{*}{\textbf{Task}}} & 
    \multicolumn{1}{c}{\multirow{2}{*}{\textbf{Model}}} & 
    \multicolumn{12}{c}{\textbf{Languages}} \\
    \multicolumn{2}{c}{\multirow{2}{*}{}} &	\multicolumn{2}{c}{\textbf{EN}}	&	\multicolumn{2}{c}{\textbf{ES}}	&	\multicolumn{2}{c}{\textbf{FR}}	&	\multicolumn{2}{c}{\textbf{ZH}}	&	\multicolumn{2}{c}{\textbf{RU}}	&	\multicolumn{2}{c}{\textbf{AR}}	\\
    \midrule
    \multirow{5}{*}{\textbf{SA}}
    & 2-LM 
    & 0.4130 & 0.0118 
    & 0.4120 & 0.0160 
    & 0.4166 & 0.0076 
    & 0.3859 & 0.0156 
    & 0.6599 & 0.0101 
    & 0.6343 & 0.0232  \\
    & 2-MT 
    & \textbf{0.4588} & 0.0092  
    & \textbf{0.4554} & 0.0053  
    & \textbf{0.4448} & 0.0158  
    & \textbf{0.4260} & 0.0070
    & \textbf{0.6935} & 0.0052  
    & \textbf{0.6864} & 0.0105  \\
    & CLM 
    & 0.3183 & 0.0099  
    & 0.3351 & 0.0198  
    & 0.3066 & 0.0192  
    & 0.3104 & 0.0135
    & 0.5693 & 0.0107  
    & 0.5886 & 0.0106  \\
    & MLM 
    & 0.3236 & 0.0270  
    & 0.3188 & 0.0188  
    & 0.3153 & 0.0088  
    & 0.2936 & 0.0107
    & 0.5434 & 0.0236  
    & 0.5804 & 0.0104  \\
    & TLM 
    & 0.2593 & 0.0298  
    & 0.2768 & 0.0589  
    & 0.2528 & 0.0487  
    & 0.2344 & 0.0539
    & 0.5537 & 0.0307  
    & 0.5487 & 0.0190  \\
    \midrule
    \multirow{5}{*}{\textbf{NER}} 
    & 2-LM 
    & 0.5830 & 0.0057  
    & 0.5616 & 0.0070  
    & 0.5627 & 0.0039  
    & 0.5653 & 0.0164
    & 0.4178 & 0.0100  
    & 0.4310 & 0.0179  \\
    & 2-MT 
    & \textbf{0.7778} & 0.0014  
    & \textbf{0.7660} & 0.0014  
    & \textbf{0.7716} & 0.0031  
    & \textbf{0.7871} & 0.0043
    & \textbf{0.6551} & 0.0088  
    & \textbf{0.7311} & 0.0099  \\
    & CLM 
    & 0.4516 & 0.0110  
    & 0.4213 & 0.0075  
    & 0.4306 & 0.0131  
    & 0.5086 & 0.0053
    & 0.3004 & 0.0034  
    & 0.3223 & 0.0054  \\
    & MLM 
    & 0.3003 & 0.0017  
    & 0.2997 & 0.0001  
    & 0.3021 & 0.0019  
    & 0.3341 & 0.0108
    & 0.2891 & 0.0001
    & 0.3094 & 0.0000  \\
    & TLM 
    & 0.3485 & 0.0074  
    & 0.3471 & 0.0152  
    & 0.3499 & 0.0173  
    & 0.4876 & 0.0230
    & 0.2941 & 0.0015  
    & 0.3094 & 0.0001  \\
    \midrule
    \multirow{5}{*}{\textbf{POS}} 
    & 2-LM 
    & 0.7241 & 0.0040  
    & 0.6607 & 0.0042  
    & 0.6848 & 0.0074  
    & 0.5964 & 0.0072
    & 0.7427 & 0.0030  
    & 0.4678 & 0.0016  \\
    & 2-MT 
    & \textbf{0.8520} & 0.0065  
    & \textbf{0.7685} & 0.0203  
    & \textbf{0.8300} & 0.0017  
    & \textbf{0.7002} & 0.0029
    & \textbf{0.8587} & 0.0055  
    & \textbf{0.6575} & 0.0032  \\
    & CLM 
    & 0.5621 & 0.0069  
    & 0.5422 & 0.0066  
    & 0.5568 & 0.0064  
    & 0.3761 & 0.0148
    & 0.4975 & 0.0140  
    & 0.3040 & 0.0106  \\
    & MLM 
    & 0.2157 & 0.0063  
    & 0.1499 & 0.0055  
    & 0.1722 & 0.0084  
    & 0.0717 & 0.0040
    & 0.1275 & 0.0080  
    & 0.1511 & 0.0127  \\
    & TLM 
    & 0.4741 & 0.0147  
    & 0.3759 & 0.0378  
    & 0.3744 & 0.0153  
    & 0.3314 & 0.0112
    & 0.3798 & 0.0097  
    & 0.2299 & 0.0215  \\
    \midrule
    \multirow{5}{*}{\textbf{NLI}} 
    & 2-LM 
    & 0.4825 & 0.0075 
    & 0.4901 & 0.0046 
    & 0.4779 & 0.0102 
    & 0.3805 & 0.0089 
    & 0.4804 & 0.0059 
    & 0.4445 & 0.0126 \\
    & 2-MT 
    & \textbf{0.6017} & 0.0105 
    & \textbf{0.5938} & 0.0119 
    & \textbf{0.5860} & 0.0087 
    & \textbf{0.5881} & 0.0031 
    & \textbf{0.5982} & 0.0025 
    & \textbf{0.5943} & 0.0053 \\
    & CLM 
    & 0.3946 & 0.0479 
    & 0.4134 & 0.0227 
    & 0.4068 & 0.0373 
    & 0.3744 & 0.0400 
    & 0.3593 & 0.0519 
    & 0.3978 & 0.0314 \\
    & MLM 
    & 0.4464 & 0.0328 
    & 0.4330 & 0.0145 
    & 0.4157 & 0.0347 
    & 0.4208 & 0.0110 
    & 0.4162 & 0.0251 
    & 0.4281 & 0.0126 \\
    & TLM 
    & 0.3063 & 0.0361 
    & 0.3573 & 0.0327 
    & 0.3940 & 0.0240 
    & 0.3122 & 0.0876 
    & 0.3892 & 0.0390 
    & 0.3360 & 0.0477 \\
    \bottomrule
    \end{tabular}
    \caption{Macro F1 score using probing technique.}
    \label{tab:F1-Probing}
\end{table*}

\begin{table*}[!ht]
    \centering
    \scriptsize
	\begin{tabular}{l@{{~~}}l c@{{\smaller$\pm$}}>{\smaller}c c@{{\smaller$\pm$}}>{\smaller}c c@{{\smaller$\pm$}}>{\smaller}c c@{{\smaller$\pm$}}>{\smaller}c c@{{\smaller$\pm$}}>{\smaller}c c@{{\smaller$\pm$}}>{\smaller}c}    \toprule
    \multicolumn{1}{c}{\multirow{2}{*}{\textbf{Task}}} & 
    \multicolumn{1}{c}{\multirow{2}{*}{\textbf{Model}}} & 
    \multicolumn{12}{c}{\textbf{Languages}} \\
    \multicolumn{2}{c}{\multirow{2}{*}{}} &	\multicolumn{2}{c}{\textbf{EN}}	&	\multicolumn{2}{c}{\textbf{ES}}	&	\multicolumn{2}{c}{\textbf{FR}}	&	\multicolumn{2}{c}{\textbf{ZH}}	&	\multicolumn{2}{c}{\textbf{RU}}	&	\multicolumn{2}{c}{\textbf{AR}}	\\
    \midrule
    \multirow{5}{*}{\textbf{SA}}
    & 2-LM 
    & 0.5213 & 0.0068  
    & 0.5254 & 0.0083  
    & 0.5244 & 0.0135  
    & 0.4739 & 0.0096
    & 0.7421 & 0.0059  
    & 0.7522 & 0.0151  \\
    & 2-MT 
    & 0.5407 & 0.0086  
    & \textbf{0.5510} & 0.0084  
    & 0.5398 & 0.0054  
    & 0.4956 & 0.0093
    & 0.7522 & 0.0056  
    & \textbf{0.7767} & 0.0156  \\
    & CLM 
    & \textbf{0.5443} & 0.0072  
    & 0.4446 & 0.2115  
    & 0.5421 & 0.0089  
    & 0.5015 & 0.0187
    & 0.7553 & 0.0015  
    & 0.5283 & 0.2328  \\
    & MLM 
    & 0.5441 & 0.0107  
    & 0.5466 & 0.0314  
    & 0.5348 & 0.0237  
    & 0.4972 & 0.0142
    & 0.7509 & 0.0135  
    & 0.5695 & 0.1427  \\
    & TLM 
    & 0.5358 & 0.0186  
    & 0.5501 & 0.0128  
    & \textbf{0.5474} & 0.0137  
    & \textbf{0.5069} & 0.0119
    & \textbf{0.7586} & 0.0057  
    & 0.4599 & 0.0943  \\
    
    \midrule
    \multirow{5}{*}{\textbf{NER}} 
    & 2-LM 
    & 0.8200 & 0.0042  
    & 0.8092 & 0.0053  
    & 0.8259 & 0.0035  
    & 0.8626 & 0.0022
    & 0.7215 & 0.0122  
    & 0.7274 & 0.0093  \\
    & 2-MT 
    & \textbf{0.8670} & 0.0017  
    & \textbf{0.8651} & 0.0022  
    & \textbf{0.8727} & 0.0018  
    & \textbf{0.8897} & 0.0042
    & \textbf{0.7934} & 0.0039  
    & \textbf{0.8685} & 0.0046  \\
    & CLM 
    & 0.7950 & 0.0064  
    & 0.8053 & 0.0028  
    & 0.8099 & 0.0044  
    & 0.8129 & 0.0021
    & 0.6622 & 0.0182  
    & 0.5994 & 0.1880  \\
    & MLM 
    & 0.8635 & 0.0123  
    & 0.8580 & 0.0142  
    & 0.8706 & 0.0055  
    & 0.8739 & 0.0199
    & 0.7629 & 0.0172  
    & 0.4113 & 0.2254  \\
    & TLM 
    & 0.7908 & 0.0028  
    & 0.8024 & 0.0081  
    & 0.8067 & 0.0047  
    & 0.8120 & 0.0032
    & 0.6758 & 0.0312  
    & 0.3094 & 0.0000  \\
    
    \midrule
    \multirow{5}{*}{\textbf{POS}} 
    & 2-LM 
    & 0.8925 & 0.0039  
    & 0.7365 & 0.0025  
    & 0.8496 & 0.0034  
    & 0.8088 & 0.0059
    & 0.8984 & 0.0055  
    & 0.7769 & 0.0102  \\
    & 2-MT 
    & \textbf{0.9314} & 0.0024  
    & 0.7826 & 0.0235  
    & \textbf{0.8866} & 0.0074  
    & \textbf{0.8842} & 0.0059
    & \textbf{0.9285} & 0.0029  
    & \textbf{0.8660} & 0.0088  \\
    & CLM 
    & 0.8752 & 0.0042  
    & 0.7854 & 0.0024  
    & 0.8573 & 0.0041  
    & 0.7906 & 0.0195
    & 0.8264 & 0.0104  
    & 0.5932 & 0.0194  \\
    & MLM 
    & 0.9177 & 0.0068  
    & \textbf{0.8079} & 0.0259  
    & 0.8851 & 0.0019  
    & 0.8313 & 0.0079
    & 0.9226 & 0.0048  
    & 0.8602 & 0.0132  \\
    & TLM 
    & 0.8782 & 0.0045  
    & 0.7830 & 0.0067  
    & 0.7421 & 0.2503  
    & 0.7876 & 0.0271
    & 0.8247 & 0.0088  
    & 0.6201 & 0.0071  \\
    \midrule
    \multirow{5}{*}{\textbf{NLI}} 
    & 2-LM 
    & 0.5771 & 0.0067 
    & 0.5760 & 0.0088 
    & 0.5658 & 0.0085 
    & 0.4766 & 0.0058 
    & 0.5629 & 0.0052 
    & 0.5350 & 0.0070 \\
    & 2-MT 
    & \textbf{0.6183} & 0.0054 
    & \textbf{0.6151} & 0.0082 
    & \textbf{0.5991} & 0.0073 
    & \textbf{0.5302} & 0.0086 
    & \textbf{0.5887} & 0.0041 
    & \textbf{0.5678} & 0.0032 \\
    & CLM 
    & 0.4240 & 0.2315 
    & 0.5589 & 0.0355 
    & 0.5493 & 0.0404 
    & 0.4729 & 0.1123 
    & 0.5507 & 0.0265 
    & 0.4554 & 0.1199 \\
    & MLM 
    & 0.5927 & 0.0189 
    & 0.5719 & 0.0487 
    & 0.5282 & 0.0964 
    & 0.4618 & 0.0453 
    & 0.5775 & 0.0069 
    & 0.5247 & 0.0221 \\
    & TLM 
    & 0.4428 & 0.1751 
    & 0.4728 & 0.1731 
    & 0.5345 & 0.1076 
    & 0.4558 & 0.0722 
    & 0.5061 & 0.0771 
    & 0.3816 & 0.1562 \\
    
    \bottomrule
    \end{tabular}
    \caption{Macro F1 score after model fine-tuning.}
    \label{tab:F1-Finetuning}
\end{table*}

In \Cref{tab:F1-Probing} and \Cref{tab:F1-Finetuning}, we present the macro-f1 score of models in the downstream evaluation.

\end{document}